\documentclass[lettersize,journal]{IEEEtran}
\usepackage{subfigure}
\usepackage{amsmath,amsfonts}

\usepackage[linesnumbered,ruled,vlined]{algorithm2e}

\usepackage{amsmath}  
\usepackage{array}
\usepackage[caption=false,font=normalsize,labelfont=sf,textfont=sf]{subfig}
\usepackage{textcomp}
\usepackage{stfloats}
\usepackage{url}
\usepackage{verbatim}
\usepackage{graphicx}
\usepackage{cite}
\usepackage{soul}
\usepackage{color}
\usepackage{booktabs}
\usepackage{threeparttable}
\usepackage{bbding}
\usepackage{pifont}
\usepackage{diagbox}
\usepackage{tabularx,booktabs}

\begin{document}

\title{QUIDS: Quality-informed Incentive-driven Multi-agent Dispatching System for Mobile Crowdsensing}

\author{Nan Zhou, 
Zuxin Li, 
Fanhang Man, 
Xuecheng Chen, 
Susu Xu, 
Fan Dang, 
Chaopeng Hong,\\
Yunhao Liu, ~\IEEEmembership{Fellow, ~IEEE},
Xiao-Ping Zhang, ~\IEEEmembership{Fellow, ~IEEE}, 
Xinlei Chen, ~\IEEEmembership{Member, ~IEEE}

\thanks{This paper was supported by Yunnan Forestry and Grassland Science and Technology Innovation Joint Special Project (grant NO. 202404CB090017), the Natural Science Foundation of China under Grant 62371269, National Key R\&D program of China (2022YFC3300703), Guangdong Innovative and Entrepreneurial Research Team Program (2021ZT09L197), and Tsinghua Shenzhen International Graduate School Cross-disciplinary Research and Innovation Fund Research Plan (JC20220011) and Meituan Academy of Robotics Shenzhen.}

\thanks{ 
A preliminary version of this article appeared in IEEE International Conference on Computer Communications (IEEE INFOCOM 2024)~\cite{li2024quest}
}

\thanks{
Nan Zhou, Zuxin Li, Fanhang Man, Xuecheng Chen are with Shenzhen International Graduate School, Tsinghua University, China. 
E-mail: \{zhoun24, lizx21, mfh21, chenxc21\}@mails.tsinghua.edu.cn
}

\thanks{
Susu Xv is with Department of Civil and System Engineering, Johns Hopkins University, United States of America.
E-mail: sxu83@jhu.edu
}

\thanks{
Fan Dang and Yunhao Liu are with the School of Software and BNRist, Tsinghua University, Beijing 100084, China. 
Email: dangfan@tsinghua.edu.cn, yunhao@greenorbs.com
}

\thanks{
Chaopeng Hong and Xiao-Ping Zhang is with Shenzhen International Graduate School, Tsinghua University, Shenzhen, China. 
E-mail: \{hongco, xiaoping.zhang\}@sz.tsinghua.edu.cn
}

\thanks{
Xinlei Chen is with the Shenzhen International Graduate School, Tsinghua University, China.
Email: chen.xinlei@sz.tsinghua.edu.cn
}

\thanks{
Nan Zhou and Zuxin Li are co-primary authors.
}

\thanks{
Corresponding authors: Xinlei Chen.
}

\thanks{
Manuscript submitted April 2025.
}

}

\markboth{IEEE Internet of Things Journal}%
{Zhou. N \MakeLowercase{\textit{et al.}}: QUIDS}

\IEEEpubid{0000--0000/00\$00.00~\copyright~2021 IEEE}

\maketitle

\begin{abstract}
This paper addresses the challenges of achieving optimal Quality of Information (QoI) in non-dedicated vehicular mobile crowdsensing (NVMCS) system, where vehicles not originally designed for sensing are leveraged to collect real-time data as they traverse urban environments. These challenges are exacerbated by the interrelated issues of sensing coverage, sensing reliability, and the inherently dynamic nature of participating vehicles.
To tackle these challenges, we propose QUIDS, a \ul{QU}ality-informed \ul{I}ncentive-driven multi-agent \ul{D}ispatching \ul{S}ystem, which ensures high sensing coverage and sensing reliability under budget constraints in NVMCS systems. 
QUIDS improves QoI by introducing a novel metric, Aggregated Sensing Quality (ASQ), designed to quantitatively capture the concept of QoI by integrating both sensing coverage and sensing reliability.
Moreover, we develop a Mutually Assisted Belief-aware Vehicle Dispatching algorithm that estimates sensing reliability and allocates monetary incentives under uncertain vehicle conditions, thereby further improving ASQ.
Evaluation using real-world data collected from a deployed NVMCS system in a metropolitan area demonstrates the effectiveness of QUIDS. The ASQ metric shows a 38\% improvement over non-dispatching scenarios and a 10\% enhancement over state-of-the-art methods. Additionally, QUIDS reduces reconstruction map errors by 39–74\% across various reconstruction algorithms, validating its efficacy in improving QoI within NVMCS systems.
Addressing the often-overlooked issue of sensing reliability in existing studies, the QUIDS system leverages non-dedicated vehicles and incorporates a quality-informed incentive-driven dispatching system to jointly optimize sensing coverage and sensing reliability. 
This enables low-cost, high-quality, and scalable urban environmental monitoring without the need for dedicated sensing infrastructure, and makes the system applicable to diverse smart-city scenarios such as traffic monitoring and environmental sensing.
\end{abstract}


\section{Introduction}

\IEEEPARstart{N}{on}-dedicated vehicular mobile crowdsensing (NVMCS) systems have emerged as a promising paradigm for collecting large volumes of spatio-temporal data~\cite{8689081}. Non-dedicated vehicular sensing platforms, such as taxis, delivery drones~\cite{xiang_reusing_2021, chenDeliverSenseEfficientDelivery2023}, and ride-sharing vehicles like Uber and Lyft, can collect data while navigating urban environments, offering cost-effective and easily maintainable solutions for NVMCS. By harnessing the collective sensing capabilities of these non-dedicated vehicles, NVMCS system supports a wide range of applications that enhance human life and inform decision-making processes, including public infrastructure management~\cite{feltenstein1999analysis}, traffic monitoring~\cite{buch2011review}, and public policy formulation~\cite{7921604}.

\IEEEpubidadjcol

One of the key challenges in NVMCS systems is ensuring optimal Quality of Information (QoI), which depends on both sensing coverage and sensing reliability. Sensing coverage refers to the spatial and temporal extent of data collection, while sensing reliability pertains to the accuracy and consistency of sensor measurements. However, non-dedicated vehicles, which prioritize fulfilling ride requests, often tend to concentrate in high-demand areas, leading to reduced sensing coverage in less populated or remote regions. In contrast, an effective NVMCS system typically requires comprehensive, city-wide data sampling to ensure sufficient spatio-temporal granularity for meaningful analysis. This imbalance in the spatial distribution of sensing resources diminishes overall sensing coverage, thereby undermining QoI. 
A seemingly straightforward solution is to dispatch vehicles to underserved areas, where sensing coverage is inadequate. However, this approach may result in a reduction in drivers' earnings, as these areas typically experience fewer ride requests. Consequently, drivers may be reluctant to accept the proposed dispatch assignments, as they would be economically disadvantaged. To mitigate this issue, it is necessary to provide additional compensatory incentives that align the drivers' economic motivations with the objectives of the sensing system. 
Moreover, the sensors deployed on non-dedicated vehicles are inherently subject to various sources of uncertainty, which can introduce significant fluctuations in sensor readings. These fluctuations arise from multiple factors, including measurement inaccuracies, sensor degradation over time, and the absence of routine calibration. Such uncertainties can severely compromise the reliability and accuracy of the sensor data, which is critical for the overall performance of the NVMCS system. 


In addition, the interplay between sensing coverage and sensing reliability often leads to a trade-off. Specifically, evenly distributing sensors across the sensing area can improve coverage; however, this may reduce the number of sensors in each sub-region, thereby increasing measurement uncertainties and diminishing reliability. Conversely, concentrating sensors in specific regions can enhance reliability by generating more data points, but this approach compromises coverage in other areas, resulting in data gaps and reduced spatial resolution.
This inherent trade-off between sensing coverage and sensing reliability presents significant challenges in achieving a balanced QoI, particularly in dynamic environments where sensor reliability is variable. These complexities further exacerbate the difficulty of ensuring high-quality data collection.


Existing solutions have made significant strides in enhancing the QoI in NVMCS systems. These solutions can be broadly categorized into two main approaches:
(1) \emph{Improving sensing coverage through vehicle dispatching}, which includes various dispatching strategies such as dynamic programming~\cite{8964368}~\cite{8712442}, and reinforcement learning~\cite{9488713}. However, these methods optimize only for sensing coverage and assume near-perfect sensor reliability. In reality, sensor reliability often exhibits significant uncertainty due to environmental interference and operational fluctuations. Neglecting sensing reliability may lead to unreliable data and misleading information.
(2) \emph{Ensuring the sensing reliability of individual sensors}, which focuses on improving sensor reliability using external references~\cite{cheng_ict_2019} or machine learning interpolation techniques~\cite{lin_calibrating_2018} to enhance the performance of low-cost sensors. While these approaches are effective under certain conditions, they frequently assume the availability of specific sensor types or external calibration references, which can limit their applicability or make them prohibitively expensive in many real-world scenarios~\cite{USPS2021}. Moreover, they often fail to account for the monetary incentives required for vehicle dispatching, a crucial factor in ensuring the feasibility of the dispatching plan, especially considering the limited budgets typically available in practical applications.


The challenges of ensuring QoI in NVMCS systems can be summarized as follows:
\emph{(C1) The inherent trade-off between sensing coverage and sensing reliability.} As discussed earlier, achieving an optimal balance between these two factors is critical; however, they often conflict, making simultaneous optimization difficult.
\emph{(C2) The difficulty of accurately estimating the sensing reliability of individual sensors.} The continuous movement and varying locations of sensors, along with sensor drift and other uncertainties, complicate the assessment of the accuracy and consistency of their measurements.
\emph{(C3) The customization of an effective monetary incentive strategy.} Limited incentive budgets and the willingness of vehicles to accept tasks under these financial constraints make the design of efficient incentive schemes particularly challenging.
Therefore, in vehicle dispatching scenarios with constrained budgets, optimizing both sensing coverage and sensing reliability becomes increasingly complex and dynamic, thus hindering the improvement of QoI.

To address these challenges, we propose QUIDS, a QUality-informed Incentive-driven multi-agent Dispatching System, specifically designed to enable dynamic, sensing-driven dispatching for NVMCS tasks. Specifically:
\emph{(S1) To tackle (C1)}: We introduce a novel metric, Aggregated Sensing Quality (ASQ), which effectively balances the trade-off between sensing coverage and sensing reliability. Building on the insight that the sensing reliability of multiple lower-quality sensors can be compensated by aggregating their readings, as outlined in~\cite{kassandros_data_2022}, ASQ combines data from these sensors to produce a measure that approximates the performance of higher-quality sensors.
\emph{(S2) To tackle (C2\&C3)}: We propose a Mutually Assisted Belief-aware Vehicle Dispatching algorithm, grounded in the core principles of truth discovery, which integrates both sensing reliability and monetary incentives in guiding dispatch decisions. In this algorithm, the collaborative sensing reliability is derived from the aggregated data readings of multiple sensors, while monetary incentives are calculated based on ride requests originating from various destinations. 
Sensors and vehicles are mutually assisted by exchanging information and resources, thereby enhancing both the reliability of the collected data and the operational efficiency of the dispatch process.
Moreover, the dispatch process itself influences both the inference of sensing reliability and the computation of incentives, promoting adaptive and optimized data collection that improves the overall QoI.
To evaluate the performance of our system, we deployed a NVMCS system involving 29 taxis over a two-month period to collect fine-grained air pollution data. The results demonstrate the effectiveness and potential of QUIDS in achieving optimal QoI in NVMCS scenarios.

To summarize, the main contributions of this paper are as follows:

\begin{itemize} 


    \item Propose a novel metric named Aggregated Sensing Quality (ASQ) to jointly optimize the inherent trade-off between sensing coverage and sensing reliability, thereby enhancing the Quality of Information (QoI); 
    
    \item Design a Mutually Assisted Belief-Aware Vehicle Dispatching Algorithm that enhances ASQ through real-time inference of sensing reliability;
    
    \item 
    Design a Monetary Incentive Mechanism, which incorporates human behavioral uncertainties into the modeling framework by introducing differentiated incentives, thereby effectively expanding the feasible region and solution space of the optimization problem.
    
\end{itemize}

The remainder of this paper is organized as follows. In Section~\ref{sec:prob_def}, we formally define the coupled problem of sensing coverage and sensing reliability. Section~\ref{sec: proposed solution} presents the proposed algorithmic framework in detail. In Section~\ref{sec: evaluation}, we conduct a comprehensive performance evaluation of the proposed method. Section~\ref{sec:discussions} discusses the generalizability and potential limitations of our approach. A review of related work is provided in Section~\ref{sec:related_work}. Finally, Section~\ref{sec:conclusion} concludes the paper.


  
  

\section{System model \& Definition}
\label{sec:prob_def}

We consider a NVMCS system that encounters challenges stemming from uneven sensing coverage and inconsistent sensing reliability.
To address this issue, we present case studies (Section \ref{subsec: backgroundandbasis}), describe the system model and dispatching parameters (Section \ref{subsec: system_model}), and detail how sensing reliability and sensing coverage are modeled as two key components of vehicular sensing performance (Section \ref{subsec: objfunc}), ultimately aiming to enhance QoI.

\subsection{Motivation Case Studies}
\label{subsec: backgroundandbasis}

\begin{figure}[t]
\setlength{\abovecaptionskip}{0.cm} 
\setlength{\belowcaptionskip}{-0.4cm} 
\setlength{\subfigcapskip}{-0.1cm}  %
\centering
    \subfigure[Deviations and variations of low-cost sensors by time.]{
        \centering
            \includegraphics[width=0.46\columnwidth]{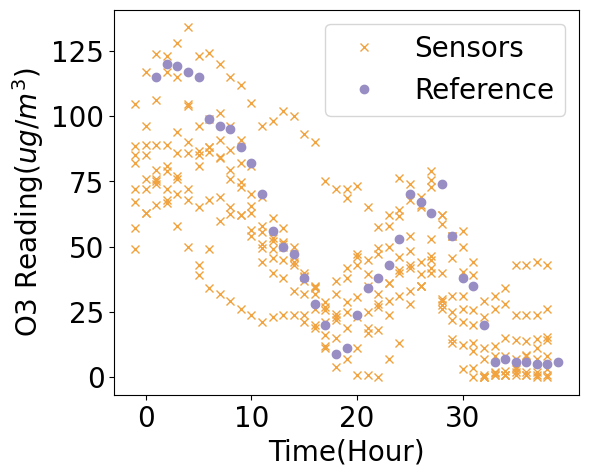}
    }
   \subfigure[Data fusions for improving sensing reliability, reproduced from~\cite{kassandros_data_2022}.]{
        \centering
            \includegraphics[width=0.46\columnwidth]{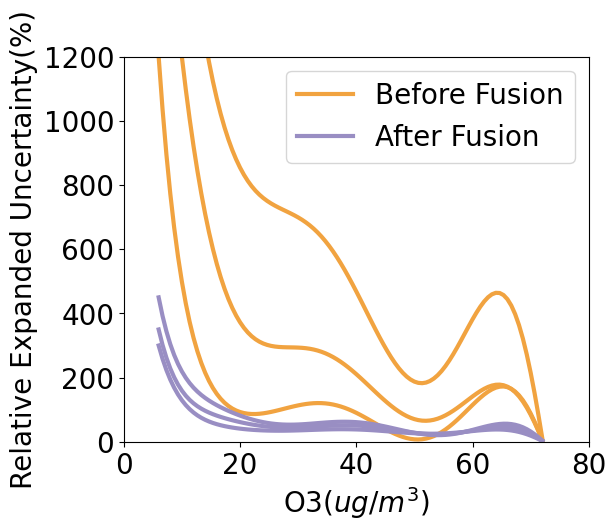}}
\caption{Motivations for QUIDS.}
\label{fig: sensor-deviation}
\end{figure}

In NVMCS systems, accurately measuring $O_3$ levels is crucial for assessing air quality and identifying potential health risks. However, widely deployed low-cost $O_3$ sensors often show substantial deviations from ground-truth measurements, as illustrated in Fig.~\ref{fig: sensor-deviation}(a). Such discrepancies compromise the reliability of the sensed data, thereby hindering informed decision-making. Furthermore, the lack of continuous calibration for sensors mounted on non-dedicated vehicles introduces additional uncertainty regarding the accuracy of their readings, further complicating the data collection process.

To address these challenges, we explore the potential of data fusion techniques to enhance sensing reliability by compensating for the limitations of low-quality sensors through increased sensor density. Prior research~\cite{kassandros_data_2022} has shown that data fusion can significantly reduce the relative expanded uncertainty among multiple sensors with substantial deviations, as illustrated in Fig.~\ref{fig: sensor-deviation}(b).
Building on this insight, the present study adopts a multi-stage approach to improve the reliability of sensor measurements in scenarios where reference monitoring data is unavailable. First, time-lagged and rolling statistical features are extracted to capture the temporal dynamics of sensor signals. Then, a Random Forest-based feature importance algorithm is applied to identify and retain the most informative features, thereby reducing the influence of noise and redundancy. Finally, machine learning models are employed to perform data fusion and reconstruction across multiple sensors, enabling the integration of complementary information from heterogeneous sources. This methodological framework is designed to enhance the stability of individual sensor outputs, expand spatial coverage, and improve the overall representativeness of the environmental data.

\subsection{System Models}
\label{subsec: system_model}

\begin{figure}
\centering
\includegraphics[width =0.5\textwidth]{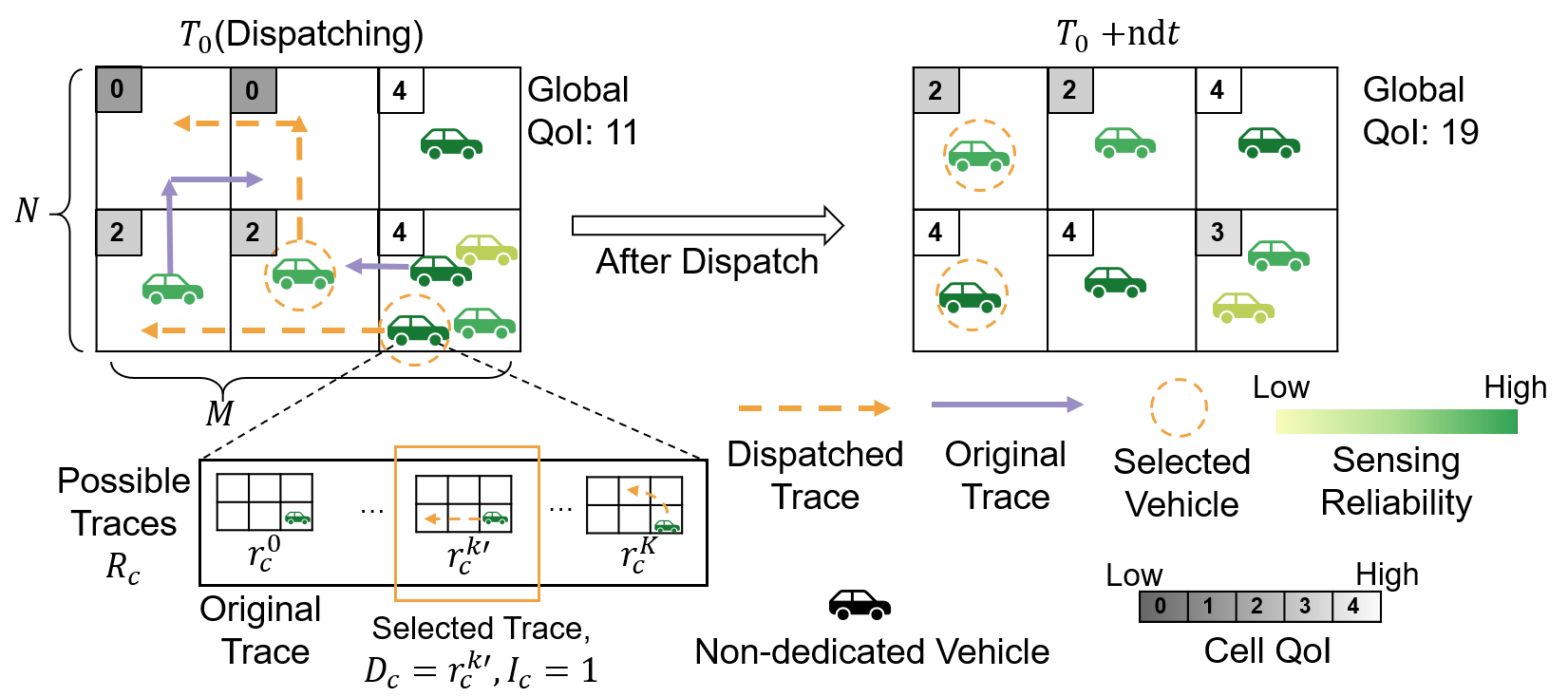}
\caption{This figure illustrates how dispatching non-dedicated
vehicular Mobile Crowdsensing (NVMCS) in uneven sensing reliability setups can improve Quality of Information (QoI) by making the coverage optimal.}
\label{fig:dispatching}
\end{figure}

The proposed dispatching system aims to enhance the Quality of Information (QoI) by optimizing sensing coverage while ensuring sensing reliability. To this end, the platform selects optimal routes for taxis to improve the system’s sensing coverage (as illustrated in Fig.~\ref{fig:dispatching}) and allocates financial incentives to drivers to promote acceptance of dispatch assignments.
While many existing dispatching systems tend to overlook the importance of sensing reliability, integrating this factor constitutes a novel contribution of our approach. We argue that an optimized spatiotemporal distribution—characterized by both balanced sensing coverage and reliable sensor performance—provides an effective and scalable solution. This design offers valuable insights for a wide range of real-world applications~\cite{10.1145/2971648.2971735, 8964368}. The key notations used throughout this paper are summarized in Table~\ref{table: signs}.

To efficiently capture and analyze the geographical area of interest, we adopt a discrete spatiotemporal representation in the form of a grid with dimensions $M \times N$. Each cell in the grid is indexed by its coordinates $(x, y)$, where $x \in {1, \ldots, M}$ and $y \in {1, \ldots, N}$ correspond to longitude and latitude, respectively. The temporal dimension is similarly discretized into time slots of fixed duration $dt$ minutes, with $t \in {1, \ldots, T}$ indexing each time slot.

Within this grid-based map, a total of $C$ vehicles are unevenly distributed and indexed by $c = 1, \ldots, C$. Each vehicle is equipped with sensors capable of automatically collecting environmental data at every time slot $t$. For each vehicle $c$, a set of candidate trajectories is defined as $R_c$, consisting of $K$ distinct routing options. Each trajectory is represented as a three-dimensional tensor $\mathbf{r}_c^k \in \mathbb{R}^{M \times N \times T}$, which encodes the vehicle’s spatial occupancy over the entire sensing period $T$. The trajectory ultimately selected for vehicle $c$ from its candidate set $R_c$ is denoted by $D_c$.

The proposed dispatch operation focuses on selecting the optimal trajectory from the set of possible traces $r_c^k \in R_c$, thereby modifying the spatiotemporal distribution of the dispatched vehicle. Each vehicle, denoted as $c$, has a default trajectory represented by $r_{c}^0$, which corresponds to the vehicle's path without any dispatch intervention. 
The vehicle trajectory $R_c$ is generated by a mobility predictor adapted from~\cite{10.1145/1869790.1869807}. Instead of operating directly on the raw road network, this method constructs a time-dependent landmark graph by mining historical trajectory data, thereby effectively encapsulating the collective driving intelligence of the driver population. Its core innovation lies in the introduction of a Variance-Entropy Clustering algorithm, which accurately quantifies the volatility and uncertainty of travel time. This enables probabilistic modeling of time-varying traffic patterns and significantly improves the accuracy of trajectory prediction. During the online computation phase, the system employs a two-stage routing strategy upon receiving a query: first, it searches the landmark graph for a sequence of landmarks to form a coarse route, and then refines this route within the actual road network to produce the final drivable path. The formula for generating candidate vehicle trajectories from historical data is as follows:
\begin{equation} 
  R_c=\text{MobilPred}(\sum_{c'=1}^C \mathcal{O}'_{c'})
\end{equation}
where $\text{MobilPred}$ is the mobility predictor, $\mathcal{O}'_{c'}\in M\times N\times T'$ represents the known historical trajectory of vehicle $c'$, $T'$ denotes the historical observation time horizon, and the set of historical vehicles $C'$ may differ from the vehicle $c$ whose trajectory is to be predicted.

To determine whether the scheduler selects vehicle $c$ and assigns it a route, we introduce an indicator variable $I_c$, defined as follows:
\begin{equation} 
I_c = \{D_c == r_{c}^k \} \in \{ 0, 1 \} 
\end{equation}

Dispatching non-dedicated vehicles may interfere with their primary missions and incur additional incentive costs. When the dispatcher decides to modify a vehicle's trajectory ($I_c = 1$), a corresponding incentive $a_c$ is provided as compensation. The proposed incentivization scheme allocates incentives to individual vehicle agents to ensure their willingness to undertake assigned tasks, while keeping the total incentive expenditure within a predefined budget constraint, denoted as $B$.
\begin{equation} \sum_{c=1}^{C} a_c \cdot I_{c} \leq B \end{equation}

Related studies in this field have investigated various compensation models aimed at incentivizing participation while minimizing disruptions to users’ routine activities~\cite{8689081, 10.1145/3185504}. In the Section~\ref{sec: evaluation}, we further examine the impact of user acceptability on the effectiveness of the proposed dispatching scheme.

\begin{table}[t]
  \centering
  \caption{Major Notations}
  \label{table: signs}
  \normalsize
  \renewcommand{\arraystretch}{1.2}
  \begin{tabularx}{\linewidth}{@{}lX@{}}
    \toprule
    Symbol & Description \\
    \midrule
    $t \in\{1, \ldots, T\}$ & $t$-th time slot for data collection \\ 
    $T$ & Number of time slots in one dispatch period \\
    $(x, y)$ & Grid coordinates ($x \in\{1, \ldots, M\}$, $y \in\{1, \ldots, N\}$) \\
    $c \in\{1, \ldots, C\}$ & The $c$-th vehicle among $C$ vehicles \\
    $I_{c}$ & Binary indicator for vehicle $c$ dispatch status \\
    $R_{c}$ & All possible trajectories for vehicle $c$ \\
    $r_{c}^k$ & The $k$-th trajectory of vehicle $c$, $r_{c}^k \in R_c, c\in\left\{0, \ldots, C\right\}$ \\
    $\mathcal{O}'_{c'}$ & The historical trajectories of vehicle $c'$ \\
    $D_{c}$ & Selected trajectory for vehicle $c$ ($M \times N \times T$ tensor) \\
    $B$ & Budget for the dispatching system \\
    $a_c \in \{1, \ldots, \mathcal{A}\}$ & Monetary incentive for sensor $c$ \\
    $B_c$ & Total monetary incentive for all vehicles \\
    $w_c \in \{1, \ldots, \mathcal{W}\}$ & Estimated sensing reliability for sensor $c$ \\
    $\beta$ & Balance factor \\
    $m_c^{(x,y,t)}$ & Reading from sensor $c$ at grid $(x,y,t)$ \\
    $m_{(*)}^{(x,y,t)}$ & Aggregated result at grid $(x,y,t)$ \\
    $b_c \in \{1, \ldots, \mathcal{B}\}$ & Constant bias for sensor $c$ \\
    $Q^{(x,y,t)}$ & Forecasted task request distribution ($M \times N \times T$ tensor) \\
    \bottomrule
  \end{tabularx}
\end{table}

\subsection{Problem Formulation}
\label{subsec: objfunc}

To balance sensing reliability and sensing coverage, we propose a novel objective function termed ASQ. The ASQ formulation is inspired by prior work in the sensing coverage domain~\cite{10.1145/2971648.2971735, 8485869}, where entropy is employed to quantify the spatial evenness of sensor distribution. We extend this concept by integrating sensing reliability into the model.
Sensing reliability is derived from the truth discovery paradigm~\cite{Meng2015TruthDO}, which characterizes sensor trustworthiness through a reliability factor $w_c$. A higher value of $w_c$ indicates a more reliable sensor, with $w_c = 1$ representing average reliability. This framework enables the identification of sensors producing more trustworthy data, as well as those that may require additional vehicle dispatching or redundant readings for compensation.
The ASQ is formulated as follows:
\begin{equation}
\phi_w(\mathcal{W}, D_c) =(1 -\beta) E(\mathcal{W}, D_c)+\beta \log Q(\mathcal{W}, D_c)
\end{equation}
where $\beta$ is a parameter that controls the relative importance of two sensing reliability-aware factors: coverage evenness and coverage rate.

The first term $E(\mathcal{W},D_c)$ represents the spatial entropy of the sensed regions. Entropy, a classic measure of uncertainty in information theory, increases as sensor coverage becomes more evenly distributed, thereby promoting fair and efficient allocation of sensing resources~\cite{schreiber2000measuring}.
The second term $log Q(\mathcal{W},D_c)$ is a coverage-quality function weighted by sensing reliability. The logarithmic transformation compresses the dynamic range of $Q(\mathcal{W},D_c)$ to match the scale of the entropy term, enhancing numerical stability and ensuring effective weighting. Moreover, the log function naturally captures diminishing returns in coverage improvement, preventing the algorithm from over-optimizing areas that already exhibit high coverage~\cite{napier1914mirifici}.

The entropy of the spatial distribution of sensed areas, denoted as $E(\mathcal{W}, D_c)$, is computed as:
\begin{equation}
E(\mathcal{W}, D_c)=-\sum_{x, y, t, \mathcal{W}} P(x, y, t, \mathcal{W}) \log P(x, y, t, \mathcal{W})
\end{equation}

The higher the entropy, the more evenly the sensors are distributed across the area, which results in better sensing coverage.
We define $P(x, y, t, \mathcal{W})$ as the aggregated sum of variance factors within the trajectory, calculated as:
\begin{equation}
\label{eqn:real_dist}
P(x, y, t, \mathcal{W})=\frac{\sum_{c=1}^{C} w_c D_{c}(x, y, t)}{CT}
\end{equation}

This equation defines the probability distribution of vehicle agents, where $D_c(x, y, t)$ indicates the presence of vehicle $c$ at location $(x, y)$ and time $t$, and $w_c$ denotes the weight associated with vehicle $c$. The sensing reliability, represented by $w_c$, acts as a weighting factor that reflects the distribution of sensing trustworthiness across the spatiotemporal domain, thereby enabling the system to prioritize reliable coverage.
The size of sensed areas, $Q(\mathcal{W}, D_c)$, is calculated simply as:
\begin{equation}
Q(\mathcal{W}, D_c) = \left|{(x,y,t): P(x,y,t,\mathcal{W}) > \frac{1}{CT}}\right|
\end{equation}
where $Q(\mathcal{W}, D_c)$ represents the size of the sensed areas in the grid map where the net sensing reliability exceeds the average sensing reliability for a sensor.

The use of a weighted aggregate sum in Eq.~(\ref{eqn:real_dist}) is motivated by its intuitive and generalizable ability to quantify sensing reliability. This formulation captures both the spatial evenness of sensor distributions and the individual reliability of each sensing agent. While the current model is designed to align with a weighted average data fusion approach, future work may investigate its integration with more advanced data fusion algorithms.
\begin{equation}
\label{eqn: overall}
\begin{aligned}
\max _{ c=1, \cdots,C  \atop k=1, \cdots, K }  \phi(\mathcal{W}, D_c) =(1 -\beta) E(\mathcal{W}, D_c)+\beta \log Q(\mathcal{W}, D_c)\\
\text{subject to \quad }
\left\{
\begin{array}{l}
 D_{c} = r_{c}^k, k \in \{ 1, 2, \ldots, K\}\\
{I_c} \in \{ 0,1\} \\
\sum_{c=1}^{C} a_c \cdot I_{c} \leq B
\end{array} 
\right.
\end{aligned}
\end{equation}

The objective function aims to maximize ASQ by balancing the entropy of sensor distributions and the extent of the sensed areas, both weighted by sensing reliability. Additionally, the model incorporates physical mobility constraints related to vehicle scheduling, along with a budgetary constraint.
This optimization problem is NP-hard, involving the combinatorial selection of discrete variables $(I_c)$ and the continuous adjustment of reliability weights $(w_c)$. It features a nonlinear objective function $\phi_w$ and a linear constraint $\sum a_c \cdot I_c \leq B$.

The novelty of this formulation lies in its simultaneous integration of sensing reliability and sensing coverage as core components of vehicular sensing performance. By embedding both factors into a unified framework, the model enables the joint optimization of sensor deployment and measurement fidelity.

\section{Algorithm Design}
\label{sec: proposed solution}

\begin{figure}
\centering
\includegraphics[width=0.95\linewidth, trim={170 0 180 0}, clip]{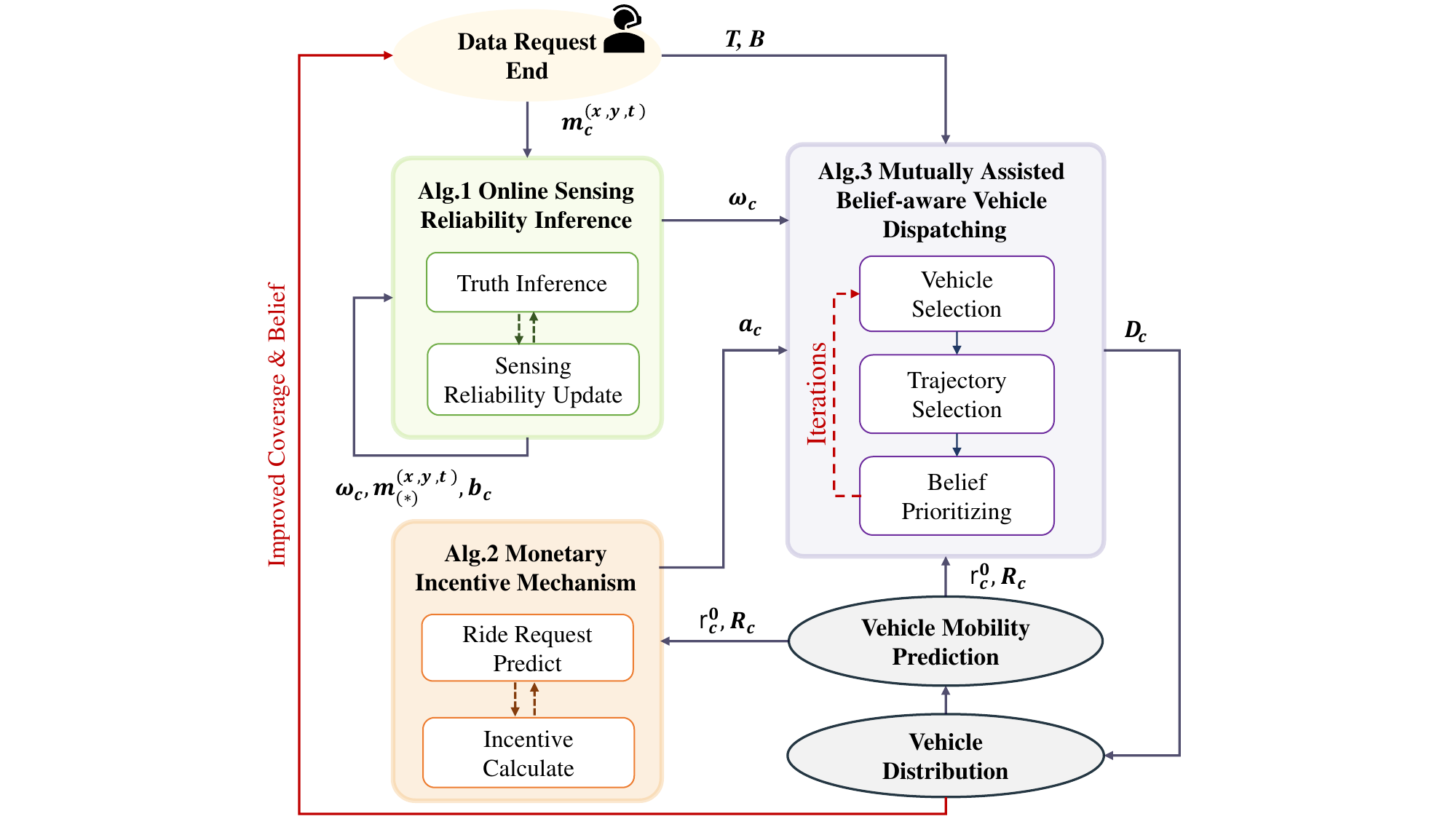}
\caption{This figure shows our proposed algorithm's framework.}
\label{fig: alg_framework}
\end{figure}

In this section, we present our proposed dispatching framework for improving QoI by optimizing ASQ. The algorithm consists of three key steps: Online Sensing Reliability Inference, Monetary Incentive Mechanism, and Mutually Assisted Belief-aware Vehicle Dispatching (shown in Fig.~\ref{fig: alg_framework}).
The first step focuses on deriving the reliability of sensors operating within a common spatial domain (Section \ref{subsec: online_dq_inferring}). The second step quantifies the monetary incentives required for vehicle dispatching (Section \ref{subsec: incen_cal}). The third step integrates the inferred sensing reliability with the calculated monetary incentives to inform the vehicle dispatching process (Section \ref{subsec: algo_parameters}).
These steps form an iterative loop, where Mutually Assisted Belief-aware Vehicle Dispatching also contributes to improvements in both Online Sensing Reliability Inference and Monetary Incentive Mechanism. Finally, we analyze the algorithm's time complexity in Section \ref{subsec: algo_analysis}.

\subsection{Online Sensing Reliability Inferring}
\label{subsec: online_dq_inferring}

In our framework, we adopt the concept of truth discovery~\cite{Meng2015TruthDO} to model the sensing reliability of sensors using the reliability factor $w_c$.
Truth discovery is a technique that infers the sensing reliability of sensors by comparing their measurements $m_c^{(x,y,t)}$ with the inferred truth $m_{(*)}^{(x,y,t)}$, derived from other sensors in correlated sensing scenarios.
However, in real-world sensing, different sensors often exhibit consistent deviations from the true value, leading to bias. This bias can significantly affect the data fusion results, particularly if all sensors dispatched to a single location share the same direction of bias, such as underestimating the true value. To address this challenge and account for systematic errors, we enhance the truth discovery approach by introducing a bias term $b_c$ and reformulating the optimization function as follows:
\begin{equation} 
\label{eqn: dq_infer_prob}
\begin{aligned} 
 \min_{\mathcal{M}, \mathcal{W}, \mathcal{B}} f\left(\mathcal{M}, \mathcal{W}, \mathcal{B} \right)=
\\
\sum_{x, y,t} \{ \sum_{c=1}^C  w_c \|m^{(x, y,t)}_{(*)} - m^{(x, y,t)}_{c} + b_c \|^2 \} \\ 
\text { s.t. } \quad \sum_{c=1}^C \exp \left(-w_c\right)=1, \sum_{c=1}^C b_c = 0
\end{aligned}
\end{equation}

In this optimization function, we aim to minimize the weighted sum of the differences between the inferred truth $m^{(x, y,t)}_{(*)}$ and the observed measurements $m^{(x, y,t)}_{c}$, while accounting for the bias terms $b_c$. The objective is to reduce the overall discrepancy between the aggregated truth and the bias-adjusted, reliability-weighted sensor readings. The first constraint limits the range of the weights to prevent them from becoming arbitrarily large or approaching negative infinity. The second constraint ensures that the bias terms do not dominate or nullify the influence of all sensor readings.

To infer sensing reliability, we formulate the optimization problem as described in (\ref{eqn: dq_infer_prob}). We adopt an approach similar to that of~\cite{Meng2015TruthDO}, employing Lagrange multipliers to solve the constrained optimization. The goal is to estimate the reliability values for each sensor accurately.
By applying Lagrangian optimization, we derive the following equations:
\begin{equation} 
\label{eqn:find_m}
\mathcal{M}: m_{(*)}^{(x,y,t)} = \frac{\sum_{c=1}^C w_c(m_c^{(x,y,t)} - b_c)}{\sum_{c=1}^C w_c}\\
\end{equation}
\begin{equation} 
\label{eqn:find_w}
\mathcal{W}: w_c = -log\frac{ \sum_{(x,y,t)} \|m_{(*)}^{(x,y,t)} - m_c^{(x,y,t)} + b_c \|^2}{ \sum_{(x,y,t)} \sum_{c^{\prime}=1}^C  \|m_{(*)}^{(x,y,t)} - m^{(x,y,t)}_{(c^{\prime})} + b_{c^{\prime}} \|^2} \\
\end{equation}
\begin{equation} 
\label{eqn:find_b}
\mathcal{B}: b_c = \frac{\sum_{x,y,t} (m_c^{(x,y,t)} - m_{(*)}^{(x,y,t)})}{\| r_c(x,y,t) \neq 0 \|}
\end{equation}

These equations facilitate the inference of sensing reliability values for individual sensors based on their measurements and the corresponding calculated weights.

Algorithm~\ref{algo:dq_inference} is designed to estimate the reliability factor $w_c$ and constant bias $b_c$ of each sensor in real-time, using the inferred truth ($m^{(x,y,t)}_{(*)}$) for all sensors over a given time period. The algorithm takes as input the sensing values, error bound $\epsilon$, and previous outputs ($m_{(*)'}^{(x,y,t)}, w_c', b_c'$), and returns the updated data quality estimates ($w_c$, $b_c$) along with the updated inferred truth ($m_{(*)}^{(x,y,t)}$).

For each spatiotemporal cell $(x,y,t)$, the algorithm iterates over all sensors within the corresponding cluster $s_{(x,y,t)}$ to update their data quality estimates ($w_c$ and $b_c$). (\ref{eqn:find_w}) and (\ref{eqn:find_b}) are employed to perform this update. Subsequently, the inferred truth is updated using (\ref{eqn:find_m}), based on the newly updated data quality estimates. These equations incorporate past outputs ($m_{(*)'}^{(x,y,t)}, w_c', b_c'$) as parameters in the summation, effectively leveraging historical data to inform the update process.

The belief in the sensing reliability inference is naturally derived from (\ref{eqn:find_w}). From this equation, we observe that the value of $w$ can be expressed as $w \propto \log k \cdot d(m_{(*)}^{(x,y,t)}, m_c^{(x,y,t)} - b_c)$, where $k$ refers to the vehicles that participate in the measurement aggregation.
Based on this, we define the belief of the estimate $\varepsilon_c$ as follows:
\begin{equation}
\label{eqn:belief}
\varepsilon_c = \log \left( \sum_{\substack{i=1 \\ i \neq c}}^C \sum_{t=1}^T \mathbf{r}_c^k(t) \cdot \mathbf{r}_i^k(t) \right)
\end{equation}

Here, $\mathbf{r}_c^k(t)$ is the $k$-th trajectory of vehicle $c$.
The product of the two tensors represents the overlap of the trajectories between all vehicles. By summing over all vehicles and time slots, we obtain the total number of vehicles that have an overlapping trajectory with vehicle $c$. Taking the logarithm of this total yields the belief signal $\varepsilon_c$. If there is no overlap between another vehicle and $c$, the belief $\varepsilon_c$ equals 0, indicating that the inferred sensing reliability for vehicle $c$ is low.

\subsection{Monetary Incentive Calculating}
\label{subsec: incen_cal}

We allocate incentives to the dispatched vehicles to ensure their willingness to perform the assigned tasks, while adhering to the budget constraints on total incentives. The utility of each vehicle is defined as its expected future returns. Given a potential scheduled trajectory $D_c^{r}$, the incentive $a_c$ is designed to compensate for the utility loss incurred by the vehicle when accepting rather than rejecting $D_c^{r}$. Since vehicles inherently seek to maximize their utility, this approach guarantees their acceptance of the proposed incentive.

\begin{algorithm}[t]
\caption{Online Sensing Reliability Inference}
\label{algo:dq_inference}
\SetAlgoLined
\SetKwInOut{Input}{Input}
\SetKwInOut{Output}{Output}
\Input{sensing value of all sensors in a given time period $m_c^{(x,y,t)}$, $c \in C$, error bound $\epsilon$, past outputs ($m_{(*)'}^{(x,y,t)}$, $w_c'$, $b_c'$)}
\Output{updated data quality estimates $w_c$, $b_c$, and inferred truth $m_{(*)}^{(x,y,t)}$}

Split the sensors into clusters based on their location, so that $s_{(x,y,t)} = \{c \mid s_c = (x,y,t)\}$\;

Initialization: Set Lagrangian factor $\lambda = 0$\;

\While{$\text{error} > \epsilon$} 
{
    $\lambda \gets \lambda + \sum_{c \in s_{(x,y,t)}} (m^{(x,y,t)}_{(*)} - m_{c}^{(x,y,t)})^2$\;

    \For{$(x,y,t) \in (M,N,T)$}
    {
        \For{$c \in s_{(x,y,t)}$}
        {
            Update $w_c$ using (\ref{eqn:find_w})\;
            Update $b_c$ using (\ref{eqn:find_b})\;
        }
        
        Update $m^{(x,y,t)}_{(*)}$ using (\ref{eqn:find_m})\;
        $\text{error} \gets |m^{(x,y,t)}_{(*)} - m^{(x,y,t)}_{(*)'}|$\;
        
        \If{$\text{error} \leq \epsilon$}
        {
            \textbf{break}\;
        }
        $m^{(x,y,t)}_{(*)'} \gets m^{(x,y,t)}_{(*)}$\;  
    }
}
\end{algorithm}

\begin{algorithm}[t]
 \SetAlgoLined
    \SetKwInOut{Input}{Input}
    \SetKwInOut{Output}{Output}
    \Input{estimated original trajectory of all vehicles $r_ {c}^0 $, a potential dispatch trajectory  $r_c^k\in R_c$, ride request predict $Q^{(x,y,t)}$, scheduling period $T$, $\{r_{min}, r_{max}\}$}
    \Output{monetary incentive $a_c$}
    Initialization: Set $a_c=r_{min}$, $t=0$;
    
    \For{$t++  \leq$ T}{
    Calculate $Q_c^0$ by \eqref{eqn:find_Q0}\;
    Calculate $Q_c^r$ by \eqref{eqn:find_QR}\;
    Calculate $a_c$ by \eqref{eqn:find_a}\;
}
\caption{Monetary Incentive Mechanism}
\label{algo: incentive}
\end{algorithm}

The core of the incentive design lies in accounting for the probability of vehicle agents receiving new task requests at their destinations. Since the primary objective of vehicle agents is to identify potential customer requests, the likelihood of obtaining ride requests at the assigned destination plays a critical role in their decision to accept a dispatched task. Therefore, the key challenge in designing the incentive mechanism is to develop a dynamic and differentiated pricing strategy under a limited budget that accurately compensates drivers for the opportunity costs and risks associated with accepting certain tasks, particularly those directed to low-demand areas. By doing so, the mechanism effectively enlarges the feasible region of the optimization model, transforming driver behavioral uncertainty into a tractable component of the solution space. This enables more flexible and efficient task assignment, ultimately maximizing both the system scheduling success rate and sensing coverage.

We adopt the model discussed in~\cite{jauhri2017space}, which predicts the number of ride requests at various locations and times within the city based on historical task request data. This prediction enables the system to effectively match ride requests with available vehicles, allowing the deployment of more vehicles within the same budget, thereby enhancing the quality of sensing coverage.

Algorithm \ref{algo: incentive} is designed to calculate the incentives required for potential dispatch trajectories within a given period. 
Let $Q^{(x,y,t)} \in [0,1]$ denote the probability that a vehicle located at a specific spatial position $(x,y)$ at time $t$ will receive at least one task request. This probability is approximated by the ratio of the number of task requests to the number of idle vehicle agents within the grid. If this ratio exceeds $1$, the task request probability is capped at $1$. 
Next, we define $r_{max}$ as the maximum monetary incentive typically accepted by the dispatch platform. 
Let the dispatching period be denoted as $T$, and define $r_u = r_{max}/T$ as the utility at each time point. Additionally, for all vehicles, the incentive has a lower bound, $r_{min}$, to ensure that the incentives provided are not negligible. 
Based on these considerations, we design the incentive $a_c$ to encourage vehicle agent $c$ to accept the potential assigned trajectory $D_c$:
\begin{equation}
\label{eqn:find_a}
a_c = \max\left( \min( r_{\max}, r_{\max} - r_u \cdot \left(Q_c^{r} - Q_c^0\right)),r_{\min}\right)
\end{equation}

\begin{equation}
\label{eqn:find_Q0}
Q_c^0 = \sum_{x,y} Q^{(x,y,t)} r_c^0
\end{equation}

\begin{equation}
\label{eqn:find_QR}
Q_c^{r} = \sum_{x,y} Q^{(x,y,t)} D_c
\end{equation}

Here, $Q_c^{0}$ represents the expected number of ride requests that vehicle $c$ can receive during period $T$ along its original trajectory $r_c^0$, while $Q_c^{r}$ represents the expected number of ride requests that $c$ can receive during $T$ along the trajectory $D_c$. The incentive $a_c$ is constrained within the range $[r_{min}, r_{max}]$. 
Based on the number of ride requests along the vehicle's original trajectory, if the scheduled trajectory $D_c$ enables vehicle agent $c$ to discover more task requests, the increase in request probability is treated as an implicit incentive, with lower monetary compensation considered. Conversely, higher compensation is provided to ensure the vehicle is willing to accept the scheduled task.

\subsection{Mutually Assisted Belief-aware Vehicle Dispatching}
\label{subsec: algo_parameters}



\begin{algorithm}[t]
 \SetAlgoLined
    \SetKwInOut{Input}{Input}
    \SetKwInOut{Output}{Output}
    \Input{estimated original trajectory of all vehicles $r_{c}^0$, possible trace set of vehicle $R_c$, dispatching budget $B$, scheduling period $T$, sensing reliability $\mathcal{W}$, monetary incentive $a_c$}
    \Output{an improved feasible solution $\mathcal{S}^{*} = \{I_c, D_c, B_c\}$}
    Initialize a feasible solution $\mathcal{S}=\{I_c, r_c^{0}, B_c\}$, set $\mathcal{S}^{*}=\mathcal{S}$, $t = 0$, belief $\varepsilon$ through \eqref{eqn:belief} \;
    \For{$t \leq T$}{
      $\mathcal{S}=\mathcal{S}^*$, $B_c=0$\;
      Calculate $P(x, y, t, W)$ by \eqref{eqn:real_dist}\;
      $C^* \leftarrow \{ c \mid \varepsilon_c = \max(\varepsilon_c) \}$ \;
      \For{$c \in C^*$}{
        $k^* = \max_{\varepsilon_c} \{ \max_r V(r_c^{k}, P) \mid r_c^{k} \in R_c \}$ \;

        \If{$k^{*}$ is the original trace}{ 
            Cancel $c$ dispatching \;
        }
        \eIf{$B_c \leq B$}{ 
          Dispatch vehicle $c$ with trace $k^*$ \;
          $B_c = B_c + a_c$
        }{
          continue \;
        }
        $c = c\rightarrow \text{next}$\;
      }
      Select $(c',k') = \arg\max_{c,k} V(r_c^{k^*}, P)$\;
      \eIf{$k' > 0$}{
        $D_c = r_c^{k^*}$ \;
        Update $\mathcal{S}^* = \{I_c, D_c, B_c\}$, $B_c$ and $\varepsilon_c$\;
      }{
        $I_{c'} = 0, B_c = 0$\;
      }
    }
\caption{Mutually Assisted Belief-aware Vehicle Dispatching}
\label{algo:schedule}
\end{algorithm}

Algorithm \ref{algo: schedule} is designed to enhance the QoI by strategically dispatching vehicles to maximize the ASQ. 
In this framework, collaborative sensing reliability is inferred through the aggregation of data from multiple sensors operating within overlapping spatial regions, while monetary incentives are computed based on the predicted ride request densities at various destinations. By exchanging information and sharing resources, sensors and vehicles mutually assist one another, thereby improving both the reliability of the collected data and the operational efficiency of the dispatching process.
Prior to initiating the scheduling procedure, the Data Request End provides the scheduling period $T$ and the monetary budget $B$. Upon completion of the scheduling, the algorithm returns the improved vehicle coverage and updated belief to the Data Request End. 
During the scheduling period $T$, the total monetary incentives allocated must not exceed the available budget $B$.

In contrast to previous work \cite{8712442}, our approach explicitly accounts for sensing reliability and its inference, rather than assuming uniform reliability across all vehicles. We prioritize vehicles with overlapping trajectories, as they are more likely to provide accurate inferred sensing reliability. These vehicles are then dispatched to less populated areas, thereby enhancing both data collection and the reliability of the inferred sensing.
To ensure that vehicles are willing to accept the scheduling, we offer incentive-based compensation, guaranteeing that their total earnings after dispatch are no less than those from executing their original trajectories.

Once the scheduling is accepted, dispatching is carried out to improve sensing coverage with respect to sensing reliability, employing a V value-based approach. This optimization enhances both overall data coverage and the quality of the collected information. The calculation of the V value is given by:

\begin{equation}
\label{eqn:vvalue}
V_c(r_c^k, P) =  - \frac{\sum_{x, y, t} w_c \cdotp r_c^k \cdotp P(x,y,t, \mathcal{W}) }{ \sum_{x,y,t}P(x,y,t, \mathcal{W})} 
\end{equation}

Here, $r_c^k$ denotes either the selected trajectory of the current vehicle or the predicted trajectory generated by the mobility predictor. The term $w_c$ represents the reliability factor, while $P(x, y, t, \mathcal{W})$ denotes the aggregated variance factor along the trajectory.

\subsection{Time Complexity Analysis}
\label{subsec: algo_analysis}

To analyze the complexity of our algorithm, we focus on the time complexity of each individual step. The initialization step has a time complexity of $\mathcal{O}(C)$, where $C$ is the number of vehicles, since each vehicle needs to be initialized. The calculation step has a time complexity of $\mathcal{O}(C T^4)$, as we need to compute the sensing quality for each pair of vehicles. The trajectory size is estimated to have a complexity of $\mathcal{O}(T^4)$, based on the use of the Bellman-Ford algorithm for trajectory optimization. Therefore, the overall time complexity of our algorithm is $\mathcal{O}(C T^4)$.

\section{Evaluation}
\label{sec: evaluation}

We present an evaluation of QUIDS through simulated dispatching and map reconstruction experiments using real-world data. 
The experimental setup uses data collected from a real-world deployment of the NVMCS system, combined with large-scale simulation-based scheduling to validate its performance (Section \ref{subsec: experiment_setup}).
We analyze how the ASQ varies with different factors and demonstrate the advantages of QUIDS over baseline approaches (Section \ref{subsec: diff_factors}). Additionally, we assess the effectiveness of the ASQ metric by exploring its relationship with downstream tasks (Section \ref{subsec: cont1_eval}). Finally, we validate the efficacy of our proposed dispatching algorithm through an ablation study (Section \ref{subsec: eval_dqinference}).

\subsection{Experiment Setup}
\label{subsec: experiment_setup}

\begin{figure}[t]
\centering
\includegraphics[width =0.47 \textwidth]{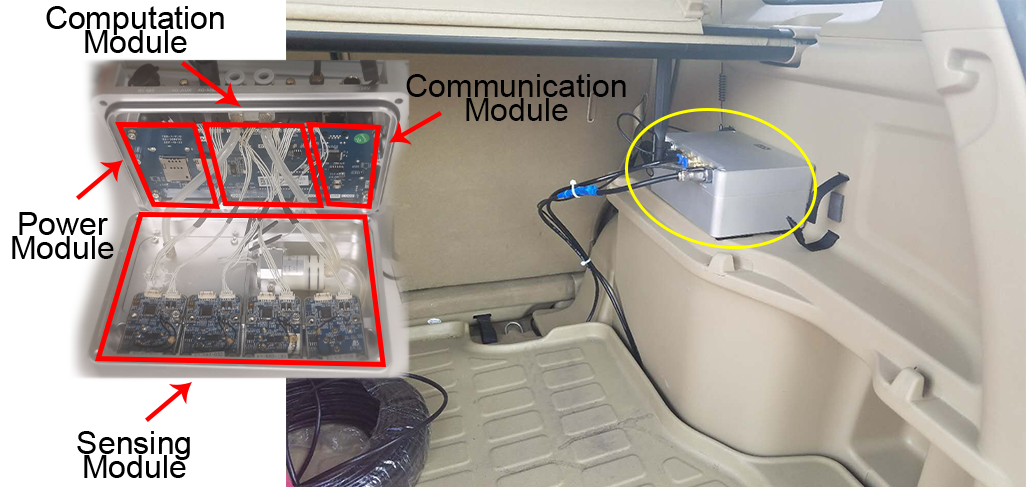}
\caption{The sensor platform deployed in our taxi, features a GPS receiver, a gas prompt, and four slacks capable of sensing various physical factors across the city.}
\label{fig: sensor-platform}
\end{figure}

\subsubsection{Real-World Data Collection and Processing}
 
We deployed mobile sensors in 29 taxis to collect data over a two-month period in a large city, capturing environmental variables such as humidity, temperature, $O_3$, and particulate matter (see Fig.~\ref{fig: sensor-platform}). Real-time GPS location data from the taxis, along with accurate sensor readings, were recorded every 3 seconds. The data underwent preprocessing, including outlier removal and imputation of missing values using a sliding window approach with a window size of 5 minutes.

\begin{figure*}[ht]
\setlength{\abovecaptionskip}{0.cm} 
\setlength{\belowcaptionskip}{-0.4cm} 
\setlength{\subfigcapskip}{-0.1cm}  
\centering
    \subfigure[ASQ vs. Budget]{
        \centering
        \includegraphics[height=4cm]{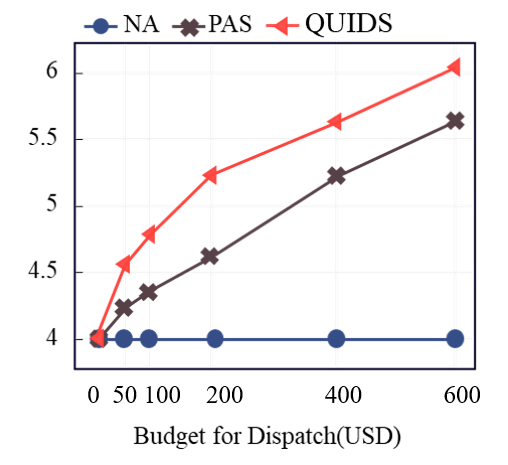}
    }
    \subfigure[ASQ vs. M. Pred. Error]{
        \centering
        \includegraphics[height=4cm]{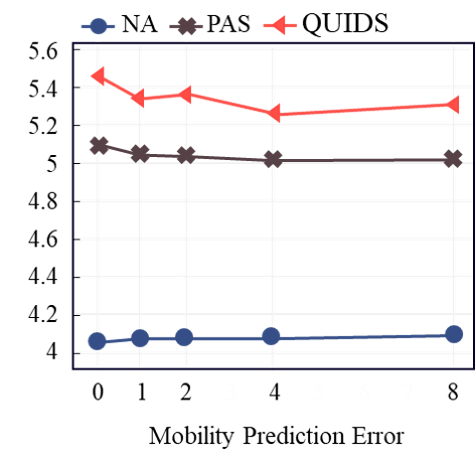}
    }
    \subfigure[ASQ vs. User Acceptance]{
        \centering
        \includegraphics[height=4cm]{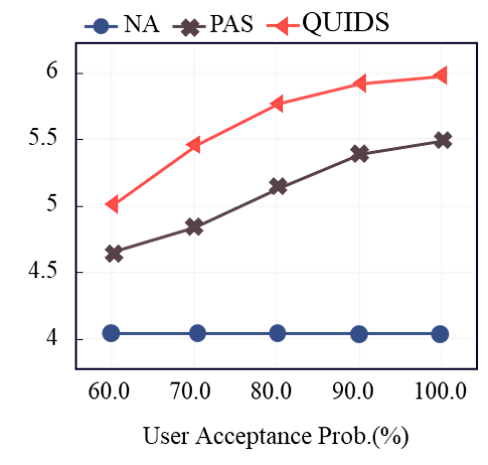}
    }
    \subfigure[ASQ vs. Error Level]{
        \centering
        \includegraphics[height=4cm]{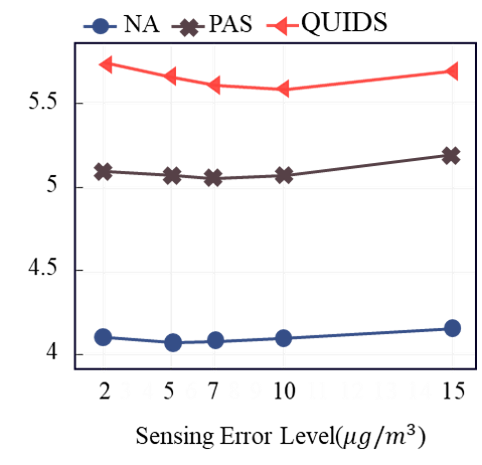}
    }
\caption{The performance of QUIDS under different factors.}
\label{fig: all_segment_map}
\end{figure*}

\begin{figure*}[ht]
\begin{minipage}[t]{1.4\columnwidth}
\centering
\includegraphics[width=1.03\columnwidth]{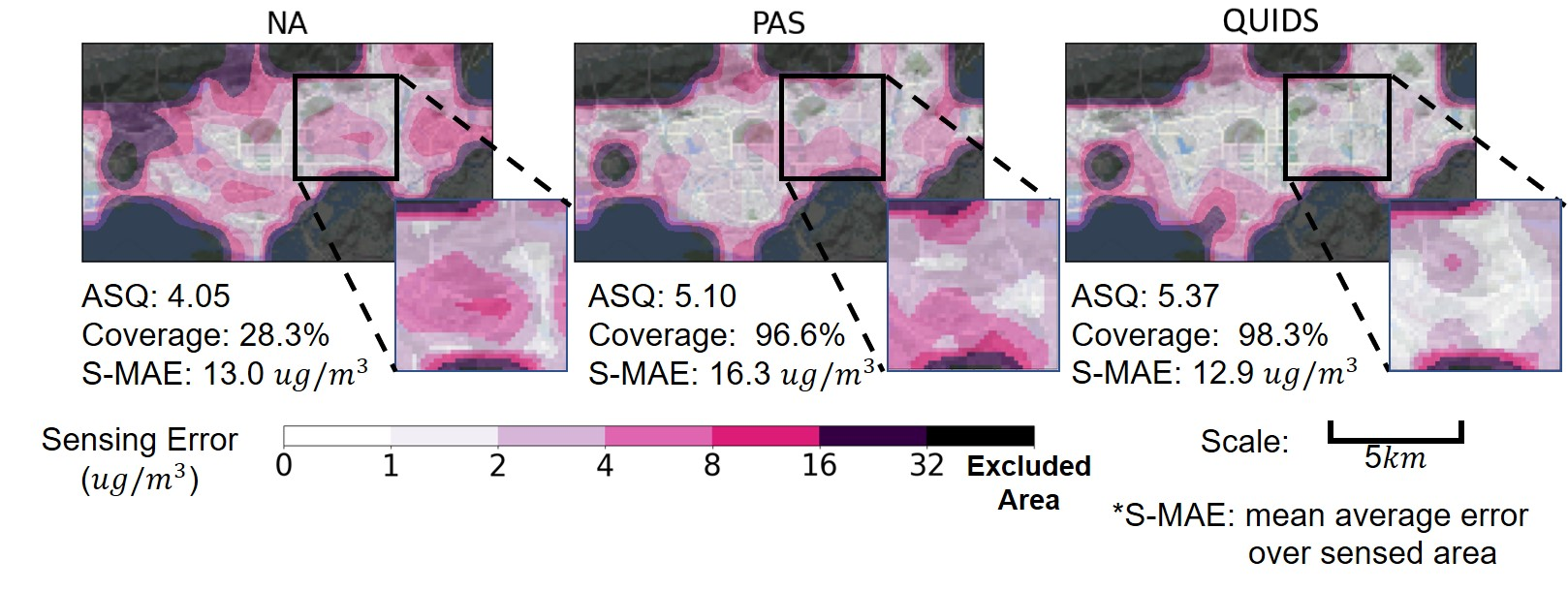}
\caption{Sensing error after dispatching. Here we zoomed in on a typical area affected by dispatching algorithms.QUIDS expanded the sensing coverage without the loss of sensing reliability.}
\label{fig: new_coverage}
\end{minipage}\hfill
\begin{minipage}[t]{0.6\columnwidth}
\centering
\includegraphics[width=\columnwidth]{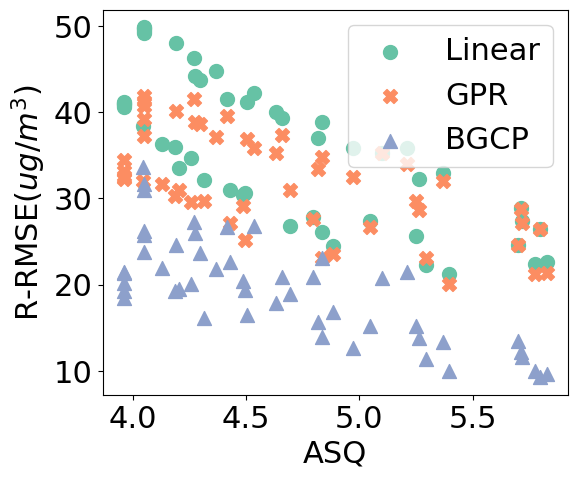}
\caption{ASQ shows negative correlations with R-RMSE, among all different reconstruction algorithms.}
\label{fig: rmse_vs_asq}
\end{minipage}%
\end{figure*}

\subsubsection{Simulation Environment Configuration}

To replicate a real-world scenario, we selected a specific area with relatively dense vehicle trajectories, corresponding to a $15\mathrm{~km} \times 8\mathrm{~km}$ grid in the city. The spatial resolution was set to $1\mathrm{~km}$, consistent with typical air pollution monitoring setups \cite{chenHAPFineGrainedDynamic2016, chenPGAPhysicsGuided2018}. The temporal resolution (i.e., time period $dt$) was set to $2$ minutes, and the actuation period was set to $5T$ ($10$ minutes), representing the average time for a taxi to travel $4\mathrm{~km}$, thereby ensuring that air quality conditions remained stable during the dispatching simulations. Due to factors such as water bodies, nature reserves, or administrative borders, 42 grid cells were not covered by any mobile sensor and were marked as excluded areas, and thus excluded from the performance metric calculations.

\subsubsection{Virtual Taxi Fleet Modeling}

To simulate a large taxi fleet and analyze their trajectories, we utilized GPS data to extract the movement patterns of each vehicle. These trajectories reflect actual taxi movements without any incentivized dispatching. To expand the simulation to include 200 virtual taxis, we adjusted the mobility patterns and spatial distribution of the original trajectories, thereby enhancing coverage. This approach allowed us to assess the behavior and sensing coverage of a larger taxi fleet without the need for additional physical vehicles.

During the dispatching process, we set the monetary budget to $B = 400$ USD, assumed zero mobility prediction error, and set the dispatch acceptance rate to 100\%. Considering the city taxi flag-down fare of 2 USD, we defined the cost parameters as $r_u = 2$ USD/min, with $r_{\min} = 2$ USD and $r_{\max} = 20$ USD. The first six weeks of data were used to train the mobility prediction and ride request models, while the remaining data were reserved for testing the proposed method.

\subsubsection{Large-Scale Sensing Simulations}

To simulate low-cost sensors with controllable sensing errors, we focus on evaluating the $O_3$ data from our dataset, as low-cost sensors for $O_3$ typically exhibit considerable variability in measurement accuracy and reliability. Given the availability of multiple relevant datasets, we leverage publicly accessible calibration data to model this variability. Specifically, we extract error distributions from established low-cost $O_3$ sensor calibration datasets~\cite{jose_m_barcelo_ordinas_2021_4570449, gonzalez_oscar_2019_3570688}, which provide detailed characterizations of the error profiles across different sensor types. These error distributions are subsequently applied to our fine-grained sensing resulting maps, ensuring that the simulated sensor types are consistent with those deployed in real-world scenarios.

To generate realistic sensor readings, we introduce sensing errors based on the extracted error distributions. These errors are systematically incorporated into the ground truth values of the sensor grid, thereby simulating the impact of low-cost sensors on both measurement accuracy and precision. By modeling sensing errors in this manner, we can rigorously evaluate the performance of our dispatching system under realistic sensing conditions, thereby closely approximating the challenges encountered in real-world low-cost sensor deployments.

\subsubsection{Baselines Methods for Comparison} Six baseline methods are adopted to evaluate the improvement in the ASQ metric achieved by QUIDS:
\begin{itemize}
    \item \textit{No Actuation (NA)}:
    This baseline refrains from any vehicle dispatch, serving as a passive reference to quantify the performance gain enabled by proactive scheduling.
    \item \textit{Prediction-Based Actuation System (PAS)}:
    A predictive incentive system that anticipates vehicle trajectories and order demand to proactively optimize sensing coverage under a limited budget~\cite{8964368}.
    \item \textit{State-Aware Hybrid Incentive Program (SHIP)}:
    A taxi dispatching scheme incorporating fine-grained vehicle state classification and a hybrid opportunity–participation incentive model to improve sensing diversity while aligning platform and driver interests~\cite{jiang2023ship}.
    \item \textit{Vehicle Assisted Data Sensing algorithm (VADS)}:
    A coalitional sensing framework based on Stackelberg game theory and Nash equilibrium optimization, designed to balance economic incentives and resource allocation across multiple operators and vehicles~\cite{zhang2024vehicle}.
    \item \textit{Vickrey–Clarke–Groves-based Mobile Sensing Tasks (VCG-MST)}:
    An enhanced auction mechanism integrating Vickrey–Clarke–Groves pricing with staggered scheduling and budget awareness to jointly guarantee passenger service quality and sensing task allocation~\cite{liu2025mechanism}.
    \item \textit{Quality-informed multi-agent dispatching system (QUEST)}: A system jointly captures sensing coverage and reliability to handle uncertain, time-varying vehicle states~\cite{li2024quest}.
    \item \textit{Graph Convolutional Cooperative Multi-Agent Reinforcement Learning (GCC-MARL)}:
    A multi-agent reinforcement learning approach using graph convolutional networks for distributed cooperative route planning, balancing passenger orders and sensing tasks ~\cite{ding2021multi}.

\end{itemize}

\subsubsection{Performance Metrics and Evaluation Protocol}

We first utilize the ASQ metric, as defined in Section \ref{subsec: objfunc}, to evaluate the performance of various dispatching algorithms. To assess the real-world impact of ASQ and these dispatching strategies, we investigate their effects through a downstream task in mobile crowdsensing: map reconstruction. Map reconstruction involves generating a comprehensive representation of environmental data across a grid map, based on the measurements collected by dispatched vehicles. For this task, we employ three distinct map reconstruction algorithms: Linear Interpolation, Gaussian Process Regression, and Bayesian Gaussian CANDECOMP/PARAFAC (BGCP)~\cite{CHEN201973}. Each of these methods provides a different approach to filling in the gaps between sensor measurements, ensuring that the reconstructed map reflects the underlying environmental conditions as accurately as possible.

To evaluate the effectiveness of the map reconstruction process, we primarily use the Reconstructed Root Mean Square Error (R-RMSE) metric. The R-RMSE quantifies the accuracy of the reconstructed map by comparing it to the ground truth. Lower R-RMSE values indicate better performance, meaning that the reconstructed map more closely aligns with the actual environmental conditions. This metric serves as a key indicator of the overall effectiveness of the dispatching strategies in terms of improving the quality and reliability of the reconstructed environmental maps.

\subsection{Evaluation for QUIDS}
\label{subsec: diff_factors}

To evaluate the potential real-world impact of various dispatching algorithms, we introduce the Error Reduction Rate (Err. Reduction), a metric that quantifies the maximum reduction in R-RMSE across all map reconstruction algorithms. This metric allows us to assess the improvements in map reconstruction accuracy achieved by each dispatching approach. Specifically, Err. Reduction represents the relative decrease in reconstruction error due to the deployment of the dispatching algorithm, with higher values indicating better algorithm performance.


Table~\ref{table:exp_results} presents a comparative analysis of the performance of various dispatching algorithms in terms of the ASQ metric and downstream field reconstruction tasks. The experimental results demonstrate that QUIDS achieves the best performance across all evaluated aspects, including the ASQ score, the Reconstructed Root Mean Square Error (R-RMSE) for three different field reconstruction algorithms, and the Error Reduction Rate (Err. Reduction). Specifically, QUIDS improves the ASQ metric by 38.02\% compared to the No Actuation (NA) method, and by 9.61\%, 5.87\%, 15.26\%, 4.88\%, and 6.48\% relative to PAS, SHIP, VADS, VCG-MST, and GCC-MARL, respectively. In terms of Err. Reduction, QUIDS achieves a 75.4\% reduction compared to NA, significantly outperforming the second-best method, SHIP, which attains a 58.67\% reduction. These results fully demonstrate the effectiveness and superiority of the QUIDS algorithm among state-of-the-art solutions. Furthermore, the differences in R-RMSE and error reduction rates across the algorithms reflect the inherent trade-offs and advantages associated with enhancing sensing coverage and improving the accuracy of map reconstruction.
  
To further assess QUIDS under varying dispatching budgets, we plot ASQ against different budget levels in Fig.~\ref{fig: all_segment_map}(a). Our proposed QUIDS consistently outperforms both baseline algorithms across all budget amounts. As the number of scheduled vehicles increases, QUIDS exhibits a more pronounced improvement compared to the PAS, achieving up to a 14.0\% enhancement with a $200$ USD budget. This improvement can be attributed to QUIDS's ability to incorporate sensing reliability, thereby allowing for more effective allocation of the dispatching budget to maximize coverage. However, as the budget increases further, the advantage of QUIDS diminishes. This trend is expected, as the distribution of vehicles becomes increasingly dense, causing QUIDS to approach its performance ceiling when most vehicles are dispatched to already covered areas.

We investigate the influence of mobility prediction errors on various dispatching algorithms by introducing random errors with varying degrees of Euclidean distance bias~\cite{huSystemLearningStatistical2006}. As shown in Fig.~\ref{fig: all_segment_map}(b), QUIDS demonstrates robustness across different levels of mobility prediction accuracy. Although ASQ decreases with increasing prediction error, QUIDS consistently outperforms the benchmark methods in most scenarios. 

In practice, some vehicles may decline incentives due to unforeseen circumstances, lack of awareness, or individual preferences. To evaluate the impact of user acceptance rates on QUIDS performance, we model the acceptance rate as the probability that each vehicle agent will accept a task after receiving the proposed incentive strategy.
We conducted 1,000 acceptance trials for each acceptance rate. Fig.~\ref{fig: all_segment_map}(c) illustrates QUIDS's performance across different time periods, with acceptance rates ranging from 60.0\% to 100.0\%. 
QUIDS consistently outperforms PAS, achieving higher ASQ scores and more effectively optimizing sensor coverage and reliability, even at lower acceptance rates.

To investigate the impact of varying degrees of sensing error on ASQ, we introduce controlled sensing errors by adding Gaussian noise, with the error level determined by the standard deviation. As illustrated in Fig.~\ref{fig: all_segment_map}(d), the relationship between sensing error levels and ASQ is complex and algorithm-dependent. In certain cases, the ASQ value remains relatively stable or even slightly increases despite an increase in sensing error variance. This phenomenon occurs because ASQ primarily reflects the relative errors between different sensors, making it less sensitive to absolute changes in the sensing error.

Fig.~\ref{fig: new_coverage} visualizes the dispatching outcomes based on ASQ. Both PAS and QUIDS dispatch vehicles from densely populated areas to sparsely populated regions. The overall sensing reliability is quantified using the Mean Absolute Error over Sensed area (S-MAE) over the sensed area. Notably, QUIDS demonstrates the lowest sensing error and the highest coverage among the evaluated algorithms, highlighting its effectiveness in balancing both sensing reliability and sensing coverage. Furthermore, an analysis of the spatial error distribution reveals that QUIDS significantly reduces errors in areas that are poorly covered by NA or PAS. This improvement can be attributed to QUIDS's strategy of selecting vehicles with higher net sensing reliability, thereby generating more accurate and reliable data compared to NA or PAS.


\begin{table*}[t]
\caption{Performance by different dispatching and reconstruction algorithms}
\label{table:exp_results}
\centering
\small 
\setlength{\tabcolsep}{4pt} 
\begin{tabularx}{\linewidth}{@{}l*{8}{>{\centering\arraybackslash}X}@{}}
\toprule
\textbf{Algorithm} & \textbf{NA} & \textbf{PAS} & \textbf{SHIP} & \textbf{VADS} & \textbf{VCG-MST} & \textbf{QUEST} & \textbf{GCC-MARL} & \textbf{QUIDS} \\
\midrule
ASQ & 4.05 & 5.10 & 5.28 & 4.85 & 5.33 & 5.37 & 5.25 & \textbf{5.59} \\
Linear R-RMSE ($\mu$g/m$^3$) & 49.72 & 28.90 & 37.67 & 25.85 & 17.41 & 26.49 & 23.84 & \textbf{24.16} \\
GPR R-RMSE ($\mu$g/m$^3$) & 39.25 & 28.81 & 27.51 & 21.59 & 13.35 & 26.48 & 19.23 & \textbf{23.93} \\ 
BGCP R-RMSE ($\mu$g/m$^3$) & 26.24 & 12.24 & 20.03 & 13.09 & 8.13 & 9.27 & 11.56 & \textbf{6.45} \\
Err. Reduction (\%) & -- & 53.32 & 58.67 & 50.12 & 69.32 & 64.6 & 55.96 & \textbf{75.4} \\
\bottomrule
\end{tabularx}
\end{table*}

\subsection{Evaluation for ASQ}
\label{subsec: cont1_eval}

Fig.~\ref{fig: rmse_vs_asq} illustrates the relationship between ASQ and R-RMSE under varying experimental conditions, including different times, budgets, and algorithms. Each point represents a single round of simulation-based dispatching. This analysis underscores the critical role of ASQ in evaluating QoI for downstream tasks, particularly map reconstruction.

The experimental results reveal a clear negative correlation between ASQ and R-RMSE: higher ASQ scores generally correspond to lower R-RMSE values, indicating higher QoI during the map reconstruction process. This consistent association confirms that ASQ effectively quantifies the core concept of QoI and translates it into a measurable system-performance indicator. Although R-RMSE may vary at the same ASQ level due to differences in reconstruction algorithms and statistical fluctuations, this does not diminish ASQ’s function as a practical bridge between the qualitative notion of QoI and its quantitative assessment in system evaluation.

\begin{table}[t]
\caption{Ablation Study}
\label{table:exp_ablation}
\centering
\small
\setlength{\tabcolsep}{4pt}
\begin{tabularx}{\linewidth}{@{}l*{3}{>{\centering\arraybackslash}X}@{}}
\toprule
\textbf{Algorithm} & \textbf{QUIDS-NoRe} & \textbf{QUIDS-NoIn} & \textbf{QUIDS} \\
\midrule
ASQ & 5.30 & 5.37 & \textbf{5.59} \\
Linear RMSE ($\mu$g/m$^3$) & 27.53 & 26.49 & \textbf{24.16} \\
GPR RMSE ($\mu$g/m$^3$) & 26.40 & 26.48 & \textbf{23.93} \\
BGCP RMSE ($\mu$g/m$^3$) & 10.58 & 9.27 & \textbf{6.45} \\
Error Reduction (\%) & 59.7 & 64.6 & \textbf{75.4} \\
\bottomrule
\end{tabularx}
\end{table}

\subsection{Ablation Study}
\label{subsec: eval_dqinference}

We conducted an ablation study to assess the contributions of Online Sensing Reliability Inference and Monetary Incentive Calculation to the overall performance of QUIDS. The configuration QUIDS-NoRe omits the Sensing Reliability Inference, where all sensors are assigned a uniform reliability level. In contrast, QUIDS-NoIn excludes the Monetary Incentive Calculation, resulting in identical incentives for all scheduled vehicles.

Table~\ref{table:exp_ablation} presents the results of this ablation study. We observe that both QUIDS-NoRe and QUIDS-NoIn achieve higher ASQ values than the baselines shown in Table~\ref{table:exp_results}; however, their performance still falls short of the full QUIDS configuration. This performance gap arises because QUIDS-NoRe does not accurately infer sensing reliability, which limits its ability to leverage optimal vehicle scheduling for error reduction. Similarly, QUIDS-NoIn fails to effectively allocate incentives, leading to a reduced number of vehicles available for scheduling. Both configurations highlight the critical importance of these two components within the QUIDS framework.

\begin{figure}[t]
\setlength{\abovecaptionskip}{0.cm} 
\setlength{\belowcaptionskip}{-0.4cm} 
\setlength{\subfigcapskip}{-0.1cm}  
\centering
    \subfigure[Inferred $w$ by QUIDS-NoRe]{
        \centering
        \includegraphics[width=0.46\columnwidth]{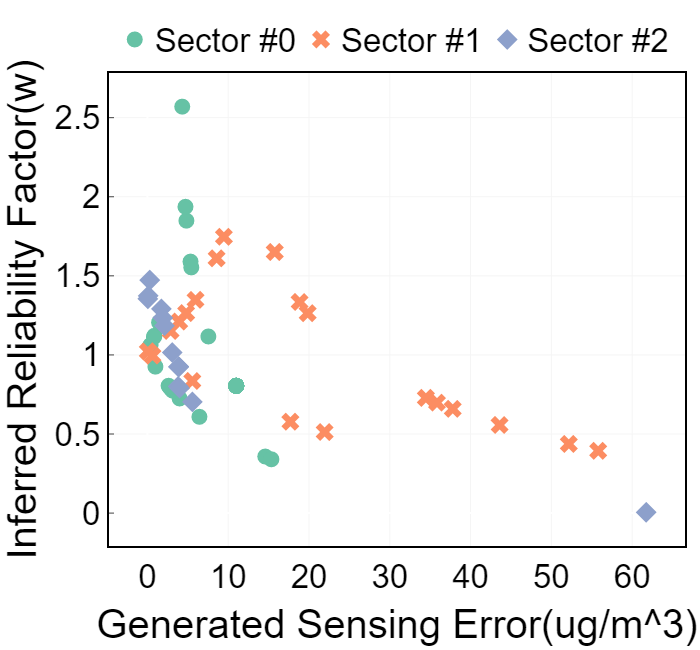}
    }
    \subfigure[Inferred $w$ by QUIDS]{
        \centering
        \includegraphics[width=0.46\columnwidth]{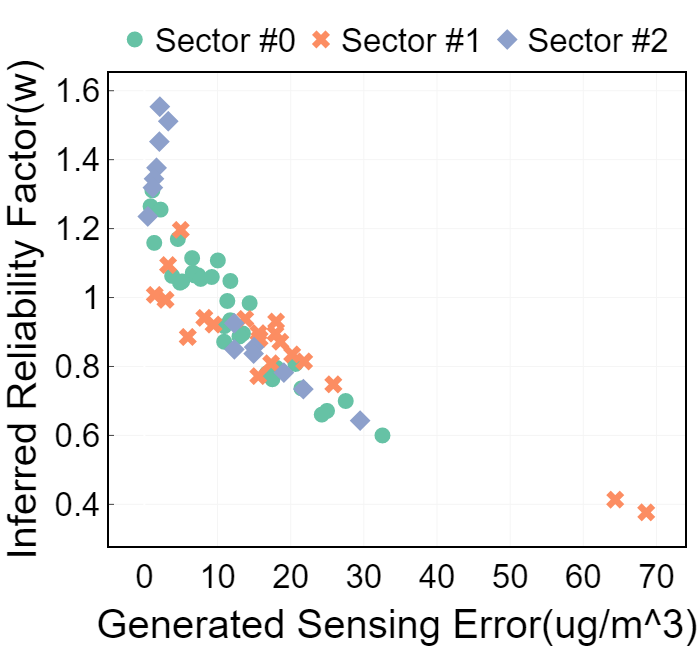}
    }
\caption{These figures visualize the relations between the generated error and inferred reliability factor. QUIDS utilizes mutually assisted dispatching to acquire a more accurate inference.}
\label{fig:sigma_w}
\end{figure}

Fig.~\ref{fig:sigma_w} illustrates the relationship between the inferred reliability factor $w$ and the resulting sensing error across different regions. For this analysis, we selected three sectors with varying numbers of vehicle agents. In Fig.~\ref{fig:sigma_w}(a), we present the results of reliability inference using QUIDS-NoRe. While the inferred sensing reliability captures some aspects of the induced errors, the local aggregation of measurements introduces biases. 
These biases can cause well-performing sensors to receive lower $w$ values due to discrepancies in the aggregated measurements. In contrast, Fig.~\ref{fig:sigma_w}(b) shows how mutually assisted dispatching enhances reliability inference. Despite minor deviations caused by inherent randomness, the results remain consistent, accurately classifying the sensing reliability of each sensor.

\section{Discussions}
\label{sec:discussions}

We discussed the potential applications and future directions of QUIDS, while also highlighting its current limitations. Specifically, these include its generalizability to other NVMCS applications (Section \ref{other NVMCS}), the potential for improving its incentive model (Section \ref{other incentive}), and its dependence on accurate mobility predictions (Section \ref{accurate}).

\subsection{Generalization to Other NVMCS Applications} 
\label{other NVMCS}
Although QUIDS was originally developed for air pollution sensing, its underlying principles are generalizable to a wide range of NVMCS tasks. For instance, QUIDS can be adapted for applications such as wireless signal sensing, noise pollution mapping, and other environmental monitoring tasks. The system's spatial granularity can be adjusted to meet the specific needs of these applications. However, since QUIDS's reliability inference model is based on truth discovery, which assumes that the data originates from the same modality, it may face challenges in scenarios that require cross-modality data fusion for sensing reliability~\cite{chenDrunkWalkCollaborativeAdaptive2015, wangCaliFormerLeveragingUnlabeled2023}. To extend QUIDS for such cases, further design modifications would be necessary to integrate and harmonize data from different sensor modalities.

\subsection{Exploring Alternative Incentive Models} 
\label{other incentive}
QUIDS currently uses a simple incentive model, but the introduction of alternative incentive strategies would not fundamentally disrupt its core contributions—modeling ASQ and implementing the mutually assisted dispatch framework. However, varying incentive patterns and scheduling methods could enhance the efficiency of vehicle dispatch and improve overall performance. For example, in cases where some vehicle agents accept tasks but do not adhere to the scheduled trajectories, two potential solutions could be explored: (1) Excluding malicious agents from the dispatch pool, and (2) Adjusting the trajectory matrix to a probabilistic model for candidates deemed potentially non-compliant, allowing for more flexible task assignment while maintaining overall reliability.

\subsection{Reliance on Accurate Mobility Predictions} 
\label{accurate}
As demonstrated in our evaluation, the performance of QUIDS is highly sensitive to the accuracy of mobility predictions. Since the algorithm relies on accurate predictions of vehicle trajectories for effective dispatch, ensuring the accuracy of mobility models is crucial for the successful real-world application of QUIDS. To mitigate the impact of prediction errors, future work could explore techniques for improving mobility prediction, such as integrating real-time data or using more advanced machine learning models for trajectory forecasting. Moreover, robustness to varying degrees of prediction error could further enhance the practical applicability of QUIDS in dynamic environments.

\section{Related Work}
\label{sec:related_work}

In this section, we review the related work in three key areas: Non-dedicated Vehicular Mobile Crowdsensing (Section~\ref{NVMCS}), Sensing Coverage (Section~\ref{SC}), and Sensing Reliability (Section~\ref{SR}).

\subsection{Non-dedicated Vehicular Mobile Crowdsensing System}
\label{NVMCS}
NVMCS leverages non-dedicated vehicles (e.g., taxis or private cars) for sensing purposes, improving both coverage and operational efficiency~\cite{rizwan2013development}. Previous research has explored various aspects of NVMCS, such as data volume~\cite{akberDataVolumeBased2018}, multi-objective trade-offs~\cite{sunMultiObjectiveOrderDispatch2023}, incentivization strategies~\cite{7065282, wangDistributedGameTheoreticalRoute2021}, and data utilization~\cite{wangSpatiotemporalTransformerData2023, chenAdaptiveHybridModelEnabled2022, chenHAPFineGrainedDynamic2016}. 
In particular, several representative approaches have been proposed in the domain of incentive strategies. For instance, the State-Aware Hybrid Incentive Program (SHIP) employs fine-grained vehicle state classification and an opportunity-participation hybrid incentive model to improve sensing diversity while aligning the interests of both platforms and drivers~\cite{jiang2023ship}. The Vehicle-Assisted Data Sensing algorithm (VADS) is a coalitional sensing framework based on Stackelberg game theory and Nash equilibrium optimization, designed to balance economic incentives and resource allocation across multiple operators and vehicles~\cite{zhang2024vehicle}. Lastly, the Vickrey–Clarke–Groves-based Mobile Sensing Task mechanism (VCG-MST) integrates VCG pricing with staggered scheduling and budget awareness to simultaneously guarantee passenger service quality and effective sensing task allocation~\cite{liu2025mechanism}.

In contrast, our study specifically addresses the challenges of sensing reliability and coverage in NVMCS. Our findings complement and extend the existing body of NVMCS research, focusing on key issues that have been largely overlooked in previous works.

\subsection{Sensing Coverage}
\label{SC}
Ensuring comprehensive sensing coverage is a critical challenge in MCS systems. Researchers have proposed spatial-temporal scheduling approaches that consider energy efficiency or budget effectiveness when selecting NVMCS agents~\cite{8528425, chenASCActuationSystem2019, renSchedulingUAVSwarm2023}. For example,~\cite{8485869} addressed spatio-temporal redundancy when performing NVMCS tasks in urban areas using high-resolution maps. Additionally,
PAS first constructs two prediction models to estimate potential vehicle routes and passenger ride-hailing probabilities across urban areas, then introduces a prediction-based execution planning algorithm to select vehicles and assign routes~\cite{8964368}.  iLOCuS remains exclusively concerned with the spatiotemporal distribution of sensed data and proposes a hierarchical iterative optimization algorithm to steer the data collected by vehicle agents toward a desired target distribution~\cite{8712442}. GCC-MARL develops a novel graph convolutional cooperative multi-agent reinforcement learning framework to achieve distributed and cooperative routing decisions, assisting taxis in balancing order-serving and sensing tasks~\cite{9488713}. All these methods optimize only for sensing coverage, operating under the assumption of nearly perfect sensor reliability. In reality, however, sensor reliability tends to exhibit significant uncertainty due to environmental disturbances and operational fluctuations. By contrast, QUIDS proposes a new metric termed Aggregated Sensing Quality (ASQ), which simultaneously captures both sensing coverage and reliability.

\subsection{Sensing Reliability}
\label{SR}
Mobile sensors are subject to environmental variations and external influences that complicate the task of ensuring reliable data collection~\cite{YOUNIS2008621}. Several methods have been proposed to address NVMCS sensing reliability, including comparing collected data with ground truth~\cite{zhang2016robust, sheng2017geometric, liuFineGrainedAirPollution2023}, calibrating sensors using external sources~\cite{cheng_ict_2019, wangHSwarmLocEfficientScheduling2023}, and leveraging machine learning models to improve sensor accuracy~\cite{lin_calibrating_2018}. However, these methods face significant challenges in dynamic, NVMCS systems, where sensor types can vary, and ground truth references are often unavailable.
Some approaches have tried to address these challenges by relying on historical data as pseudo-ground truth for static environments~\cite{zhang2014exploring, luoFieldReconstructionBasedNonRendezvous2023}. Additionally,~\cite{Meng2015TruthDO} proposed truth discovery techniques for correlated sensors and regions. While these methods provide useful insights, they remain constrained by two major issues: (1) Uneven distribution of non-dedicated vehicles: in areas with few vehicles, data coverage is sparse, resulting in unreliable sensing estimates and incomplete spatial coverage. (2) Failure to identify constant biases: existing methods primarily focus on detecting unreliable sensors but do not distinguish between dynamic errors and constant biases in sensor measurements. These limitations highlight the need for novel techniques capable of guiding dispatch decisions and enhancing the overall QoI in NVMCS systems.

\section{Conclusion}
\label{sec:conclusion}


To enhance the QoI in NVMCS systems, we propose QUIDS, a QUality-informed Incentive-driven multi-agent Dispatching System. We model the paradoxical relationship between sensing reliability and sensing coverage as an optimization problem. Our framework infers sensor reliability and calculates monetary incentives to dispatch vehicles in NVMCS systems, aiming to maximize the ASQ metric and, consequently, achieve optimal QoI. City-scale simulations based on physical features demonstrate significantly lower error rates at high coverage levels, validating the effectiveness of our approach. This solution opens new research directions for NVMCS systems, including the modeling and optimization of sensing reliability and sensing coverage under conditions of limited incentives and uncertain environments.

\bibliographystyle{IEEEtran}
\bibliography{infocom}

\vfill

\end{document}